 \documentclass[final,5p,times,twocolumn,authoryear]{elsarticle}

\usepackage{amssymb}
\usepackage{CJKutf8}
\usepackage{lipsum}
\usepackage{subcaption}
\usepackage{tabularx}
\usepackage{xcolor}

\journal{arxiv}

\begin{document}

\begin{frontmatter}



\title{L(u)PIN: LLM-based Political Ideology Nowcasting}


\author[first]{Ken Kato}
\affiliation[first]{organization={Divison of Engineering Science, University of Toronto}}
\author[second]{Annabelle Purnomo}
\affiliation[second]{organization={Department of Linguistics, University of British Columbia}}
\author[third]{Christopher Cochrane$^*$}
\affiliation[third]{organization={Department of Political Science, University of Toronto Scarborough}}
\author[forth]{Raeid Saqur$^*$}
\affiliation[forth]{organization={Vector Institute, Department of Computer Science, University of Toronto}}

\begin{abstract}
AI and LLMs have seen a drastic development over the past few years. Powerful LLMs such as GPT-4 by OpenAI became available to the public, irreversibly changing the world in various fields. However, the application of LLMs in the field of quantitative political science remains limited. At the time of writing this paper, we are only aware of a few papers making use of LLMs to analyze political speech data.

%background
The quantitative analysis of political ideological positions is a difficult task. In the past, various literature focused on parliamentary voting data of politicians, party manifestos and parliamentary speech to estimate political disagreement and polarization in various political systems. 
However previous methods of quantitative political analysis suffered from a common challenge which was the amount of data available for analysis. Also previous methods frequently focused on a more general analysis of politics such as overall polarization of the parliament or party-wide political ideological positions.

%statement of purpose
In this paper, we present a method to analyze ideological positions of individual parliamentary representatives by leveraging the latent knowledge of LLMs. The method allows us to evaluate the stance of politicians on an axis of our choice allowing us to flexibly measure the stance of politicians in regards to a topic/controversy of our choice.  

%Methods
We achieve this by using a fine-tuned BERT classifier to extract the opinion-based sentences from the speeches of representatives and projecting the average BERT embeddings for each representative on a pair of reference seeds. These reference seeds are either manually chosen representatives known to have opposing views on a particular topic or they are generated sentences which where created using the GPT-4 model of OpenAI. We created the sentences by prompting the GPT-4 model to generate a speech that would come from a politician defending a particular position. 
%validation
We validate the obtained results in two ways. First, we analyze the semantics of different projection segments by using the topic modelling technique of BERTopic to ensure that the segments are aligned with what we would expect the segments to represent. Second, we compare the obtained projections against expert estimations from Mielka, a NPO in Japan aiming to make Japanese politics more transparent to the public. They have looked at official party manifestos and policy summaries to produce an estimate of how the parties are positioned in regards to various topics. 

We find that our method yields an estimation of ideological positions that align with expert estimations from Mielka with significantly less human intervention. We show that LLMs are a great tool to empower political scientists aiming to quantitatively analyze political ideologies of representatives.

\end{abstract}




\end{frontmatter}

%\tableofcontents



\section{Introduction}
\def\thefootnote{*}\footnotetext{These authors contributed equally to this work}
Estimation and comparison of ideological position of politicians is a difficult task. Previously, political scientists have looked at roll-call voting data and word-embedding techniques to analyze individual or party-wide relative ideological positions. \citep{Word-embeddings-for-analysis-of-ideological-placement,spatial-model-for-legislative-roll-call} However, the previous papers often have the following two limitations: 
\begin{itemize}
    \item There is not enough speech data per individual representative usually to learn a latent representation for individual politicians.
    \item In some parliaments, not all roll-vote data is collected and recorded and instead, only the final result of whether a bill was passed or rejected is recorded. This limits the amount of data available to create representations for individual politicians.
\end{itemize}
Instead of learning representations for individual representatives from parliamentary speeches or looking at roll-call voting data which is limited for the Japanese diet, this paper uses the latent knowledge of LLMs to create representations for individual representatives based on their parliamentary speeches. LLMs are AI models that are trained on learning the relationships between words. During the training phase, the LLM tries to predict the next word token given a window of previous word tokens and adjusting the parameters using the back-propagation method. After the training phase, the LLM is able to embed words to vectors in a meaningful manner and these embeddings can be used to embed entire documents consisting of multiple words.\citep{reimers2019sentencebert} LLMs can also be used to classify sentences. For our research, we have fine-tuned a BERT classifier originally created by Tohoku University\citep{Tohoku} using our own self-labelled dataset to classify opinion-based sentences. This allows us to apply a filter to the large amount of speech data available for each representative and to create a more refined set of speeches to analyze.

Our method first collects the speeches of politicians for various topics using multiple query words for each topic. We then proceed to classify speech segments using our fine-tuned BERT classifier to extract the opinion-based sentences for each politician for each topic. Then we embed all of the opinion-based sentences and create an average embedding for each politician to create an ideological representation of the representative in regard to various topics of interest. This ideological embedding vector is then collapsed onto a reference axis that is created by either taking the difference of manually picked representatives or the difference of generated speeches defending a particular position.

To measure the validity of our research, we compared the scalar values obtained after collapsing the representatives onto the reference axis with expert estimations. Namely, we looked at the evaluation of the party positions by Mielka which is a NPO operating in Japan aiming to make Japanese politics more transparent to the public. We find that the scalar evaluation of the ideological positions obtained with our method aligns with the expert estimations from Mielka, verifying that our method is able to yield valid estimations with significantly less manual human intervention. 

We also looked at the projection of the ideological embedding in a lower dimension. Using the UMAP package\citep{mcinnes2018umap-software} we can reduce the dimensions of the embedding from 768 to 2 while minimizing information loss and generate plots for the reduced dimensions on a 2D plane. This allows us to analyze the relative ideological positions of politicians and the generated reference sentences. 

We believe that our method is a good reference for political scientists looking to use LLMs in their own works to obtain quantitative analysis of ideological positions.

% - Context of measuring ideological polarization in party systems 
% - Context of word and document embeddings, neural embeddings 
% - Context of Japanese politics

% - Our method allows us to express the stance of politicians on the division/polarization of our choice(left right, internationalist nationalist, valence

\section{Related Work}
In this section we present a variety of related works ranging from earlier methods to quantitatively express political ideological positions to modern methods using neural networks and LLMs. We also cover a range of works that analyze parliamentary speech data in general.

\subsection{Earlier methods:Roll call analysis}

The Dynamically Weighted Nominal Three-step Estimation (DW-NOMINATE) scores developed by \citeauthor{dw-nominate} remains the benchmark to compare the spatial orientation of the legislator ideologies until today. It is the extension of the NOMINATE score \citep{spatial-model-for-legislative-roll-call} developed by the same authors a few years earlier and they are based on the roll-call votes of the representatives. Their algorithm aims to estimate the parameters of a utility function for each representative and expresses each representative as a point on the Euclidean space. Their work laid the foundation of spatial expression of legislator ideologies and is used as a gold-standard reference for political scientists today.

There are however limitations with their approach. The method depends heavily on the availability of the roll-call votes data which in the case for Japanese politics, is only available for a limited number of bills. It also places the politicians on a left-right singular dimension which might not be appropriate for all democracies. As \citeauthor{dw-nominate} explain, in the US political system issues can be 'bundled' together. For example, a legislator who opposes gun-regulation most likely is pro-life in regards to abortion and likely is against affirmative action. This interpretation of politics is however not applicable to all democracies. In Japanese politics, knowing that a politician is against the acknowledgement of the Japanese Self-Defence Force(JSDF) in the Japanese constitution does not always tell you how the politician stands in regards to other topics. Therefore, the DW-NOMINATE score cannot be used when we are seeking a more granular analysis of political stance. 

\subsection{Earlier methods:Word frequency based approach}
\subsubsection{Wordscores - \citealt{LAVER_BENOIT_GARRY_2003}}
One of the earlier methods utilizing text data to position political parties was developed by \citeauthor{LAVER_BENOIT_GARRY_2003}. Their method was one of the first to treat text data not as a discourse to be read, interpreted and understood but as a collection of word data. Their method focused specifically on the frequency of the words in documents where the political positions were known (reference texts) and documents where political positions were unknown.(virgin texts) Their method used a 3 step process where: 1) They collected reference texts of which the political positions were known; 2) They generated word scores which represented the probability that we are reading a specific reference text given only that word; 3) Score the virgin texts using the word scores generated.

In their work, they show that the position estimations for British and Irish parties they obtain from the party manifestos are in agreement with previous position estimations and they export this methodology to the German political system as well. Furthermore, they apply the word scores they obtained from the reference party manifestos to legislative speeches to estimate the positions of individual Irish politicians as well as parties. 

\subsubsection{Wordfish - \citealt{wordfishing}}
Another early word-frequency based approach is the wordfishing approach developed by \citeauthor{wordfishing} which is still widely used among political scientists to estimate policy positions from text data. It was created to overcome the limitations of the Wordscores approach such as: 1) The need for reference texts that can be difficult to be agreed upon among political scientists; 2) Its limited applicability to time-series estimation as the political lexicon is constantly changing. 
To overcome the aforementioned limitations of the Wordscore approach, the authors use a statistical model of word counts by assuming a Poisson distribution for word frequencies. This allows them to estimate the policy positions of text data without having to prepare anchor/reference texts. 

Their method has been used by various authors such as \citeauthor{sharks-minnows},\citeauthor{Curini_Hino_Osaka_2020} and \citeauthor{CATALINAC_2018} and has allowed political scientists to analyze policy positions from text data in various forms and languages.

However, recent works have pointed out the limitations of the word frequency based approaches. For example, \citeauthor{Political-Text-Scaling-Meets-Computational-Semantics} point out that using frequency based approaches for text analysis, semantically similar words can be treated as dissimilar words hence semantically similar sentences being treated as dissimilar sentences.

\subsection{Embedding-based methods}
In the past, researchers have used embedding based methods to analyze and spatially position legislative speeches. For example, \citeauthor{Debating-Evil:-Using-Word-Embeddings-to-Analyze-Parliamentary-Debates} used the Word2Vec algorithm developed by Google to analyze legislative speeches that deal with war criminals and Nazi collaborators in post-war Netherlands. They analyzed how the conversations evolved from focusing on the crimes these individuals had committed against the nation to the suffering of the victims during the time period. They also analyzed how the conversations shifted from focusing on death penalties to focusing on imprisonment. In their work, \citeauthor{Debating-Evil:-Using-Word-Embeddings-to-Analyze-Parliamentary-Debates} come up with a workaround to use multiple word2vec models trained on different corpora to conduct a diachronic analysis between different time periods. 

Another example of a text embedding-based approach to political positioning would the the work of \citeauthor{Political-Text-Scaling-Meets-Computational-Semantics}. In their work, they combined a symbolic text representation such as tf-idf or a semantic text representation such as word embeddings with an unsupervised, graph-based clustering algorithm such as PageRank or HFLP. Through this method, they challenge the traditional word frequency-based approaches discussed earlier and validate that their method is able to position political text better than the frequency-based methods in multiple languages.

\citeauthor{Embeddings-Based-Clustering-for-Target-Specific-Stances} use a different algorithm to embedding political text. They use Google's Convolutional Neural Network (CNN) based multilingual text encoder to analyze the tweets of the Turkish public during the presidential and parliamental election of 2018 in regards to various topics. They undergo a multi-step process of manual labelling, label propagation, embedding and clustering to reveal significant degree of polarization of the public in regards to topics such as Trump, Syria and Erdogan. They also reveal the correlation across topics by looking at the mutual information metric between the clusters for different topics.

Instead of focusing on the embeddings of the text data such as legislative speech or party manifesto, \citeauthor{Word-embeddings-for-analysis-of-ideological-placement} focuse on the embeddings of the indicator variables which represent the party affiliation of the text. They call this the party embeddings. In their work, they implement a shallow neural network consisting of one hidden layer that aims to predict the word at position \textit{t} given words appearing in a window $\Delta$ before and after the word. The model input consists of these surrounding words as well as the party indicator variable. After fitting the model to the parliamentary speech corpora, they got a model that is capable of embedding the party indicator variables in a meaningful manner. They then reduce the dimension of the embeddings using the PCA method and analyze the what the dimensions semantically represent. They also verify that the political positioning they retrieved are aligned with the previous methods such as DW-NOMINATE and other expert surveys. 

\subsection{Use of LLMs}
At the time of writing this paper, we are only aware of a few works that explore the application of LLMs in political ideology positioning. This section will introduce these works and their contributions. 

\citeauthor{llm-latent-position-of-politicians} use a pair-wise comparison approach of senators to position them on a liberal-conservative scale or on topics such as abortion and gun-control. In their method, they provide the GPT-3.5 model by OpenAI with two senators to compare and prompt it to give the senator who is more pro-choice, conservative or supporting of gun-control. By repeating this match up between senators, they are able to position the senators on a spectrum using the Bradley-Terry model. They show that while the \textbf{LaMP}(\textbf{La}nguage \textbf{M}odel \textbf{P}airwise comparison) scores obtained highly correlate with the DW-NOMINATE scores, the scores do not simply parrot the DW-NOMINATE scores. They argue that as LaMP scores are more aligned with surveyed political scientists as LaMP scores do not solely rely on roll-call votes like DW-NOMINATE scores do, but take into account how the politicians are depicted on main stream media and other sources that were used to train the GPT-3.5 model. 

In their work, \citeauthor{Scaling-Political-Texts-with-ChatGPT} use a prompt engineering approach to spatially position British parties and US senators on a left-right scale. They feed in British party manifestos or tweets from US senators into the GPT-4 model using the API service provided by OpenAI and task the LLM to score the given text from 0 to 100 on a left-right scale. They verify that the scores they obtained using this method correlate extensively with existing expert estimates as well as past research. The authors mention the benefit of their methodology over the work of \citeauthor{llm-latent-position-of-politicians} as their method does not require the LLM to 'know' the political actors to estimate their political ideological position as they are provided the tweets/manifestos of the political actors.

So far, the works introduced have mainly focused on one source of text data such as tweets, parliamentary speeches or manifestos. \citeauthor{Understanding-Politics-via-Contextualized} extends this approach by suggesting a framework to combine political text from multiple sources. They suggest a unique neural architecture, "Compositional Reader", capable of processing multiple sources of text in one shot and successfully training it on two major tasks of Authorship prediction and Referenced Entity Prediction. They show that this architecture is able to outperform the previous BERT-based architectures. They also show that the representations of politicians learned by the model is capable of capturing a greater depth of political context compared to previous methodologies as their representation do not simply capture the linguistic context but a broader political context.
% \begin{itemize}
%     \item risks study: \citep{LLMs-and-political-science}
% \end{itemize}

% \subsection{Other}
% \begin{itemize}
%     \item social groups polarization:\citep{Quantifying-social-organization-and-political-polarization-in-online-platforms}
%     \item Overview paper on computational text analysis: \citep{Large-Scale-Computerized-Text}
%     \item overview of quant text analysis in Japanese: \citep{quantitative-text-analysis-in-japanese}
%     \item advisory piece about embedding models in context of social science: \citep{Word-embeddings:-What-works}
%      \item measuring polarization: \citep{gentzkow-measuring-group-differences}
%     \begin{itemize}
%         \item measuring polarization over the years
%         \item measured by the ease with which an observer who know the model could guess a speakers party solely by their speech
%         \item the number of observed speeches are significantly small compared to the actual possible speeches
%         \item this introduces finite sample bias where the number of possible choices are significantly larger compared to observerd phenomenon
%         \item they use statistical estimators to address this (leave out estimator and maximum likelihood estimator
%         \item cross-validate against previous measures of polarization 
%     \end{itemize}
    
% \end{itemize}

\section{Parliamentary Speech Data}
% \begin{itemize}
%     \item - Description of the dataset 
%    - Parties 
%    - How many politicians are there? 
%    - How many politicians in each party
%    - Year
%    - etc. to contextualize the reader
%    \item - Cleaning 
%     - process through BERT classifier 
% \end{itemize}

In this study, we utilized the Japanese government's API to acquire parliamentary speech data, specifically focusing on the main political parties in Japan, namely the Liberal Democratic Party, National Democratic Party, Constitutional Democratic Party, Japanese Communist Party, Komeito, and Japan Restoration Party, each with a membership of more than four parliamentary representatives.

Our analysis centered on a total of 464 representatives in the Japanese lower house, who were actively serving as of May 2023. To narrow our focus, we collected speeches pertaining to key topics of Defense and Nuclear Power. To facilitate this process, we defined multiple keywords for each topic to query the speeches API. These query words are summarized in tables \ref{tab:English-query} and \ref{tab:Japanese-query} in the appendix. 

Employing a refined approach, we fine-tuned a BERT classifier model based on the cl-tohoku/bert-base-japanese-v3 \citep{Tohoku}
model, utilizing a self-labeled dataset comprising 1,439 speech segments. The classifier was specifically designed to identify various sentence types, including opinion-based sentences, fact-based sentences, questions, descriptive sentences, and others. However, for the purpose of our analysis, we exclusively extracted opinion-based sentences to create a more nuanced and focused dataset for further examination.

\section{Methodology}
\begin{figure}[h]
\centering
  \centering
  \includegraphics[width=1\linewidth]{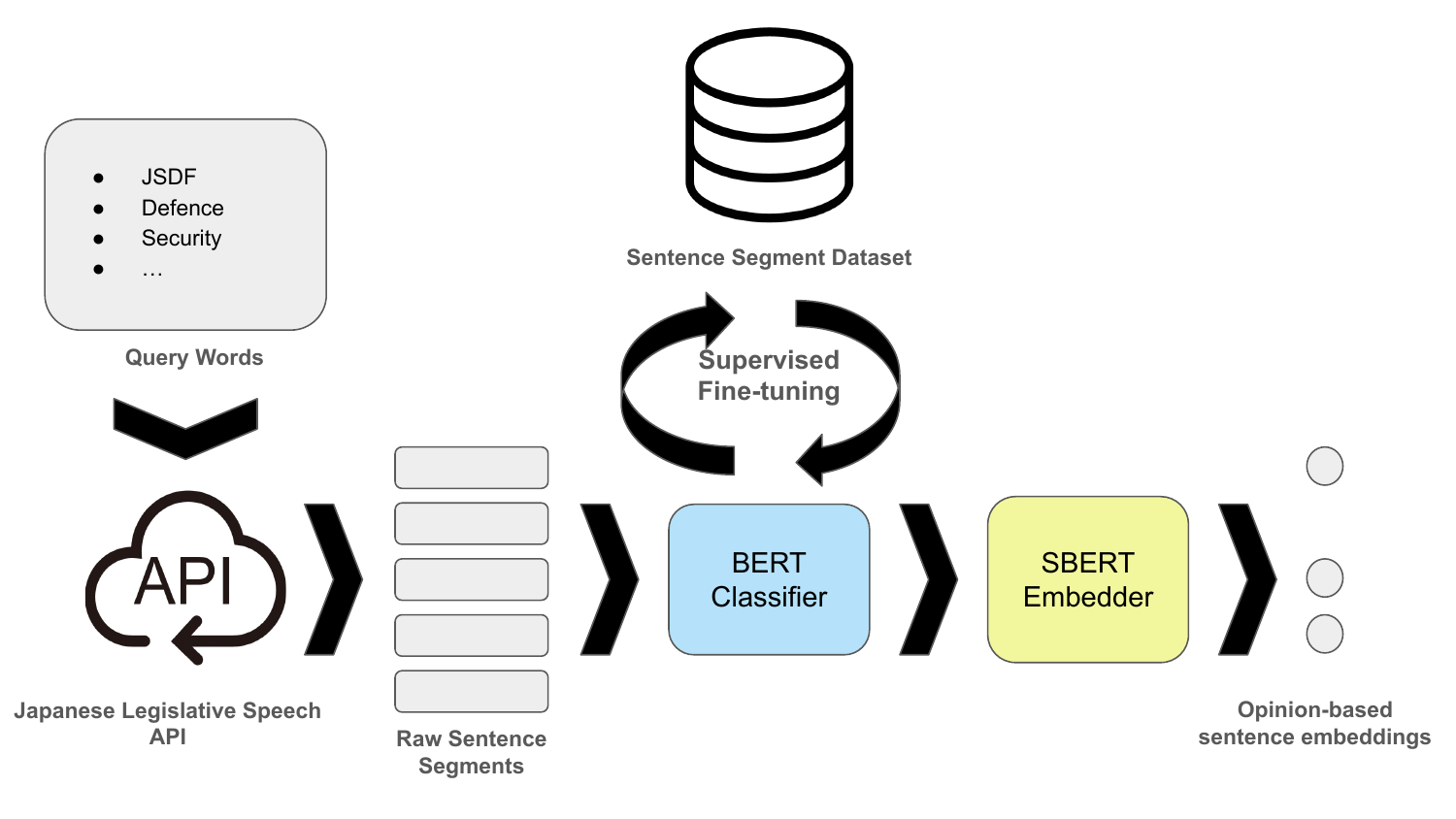}
  \caption{Graphical representation of the embedding phase of the parliamentary speeches}
  \label{fig:embeddingphase}
\end{figure}

\begin{figure}[h]
\centering
  \centering
  \includegraphics[width=1\linewidth]{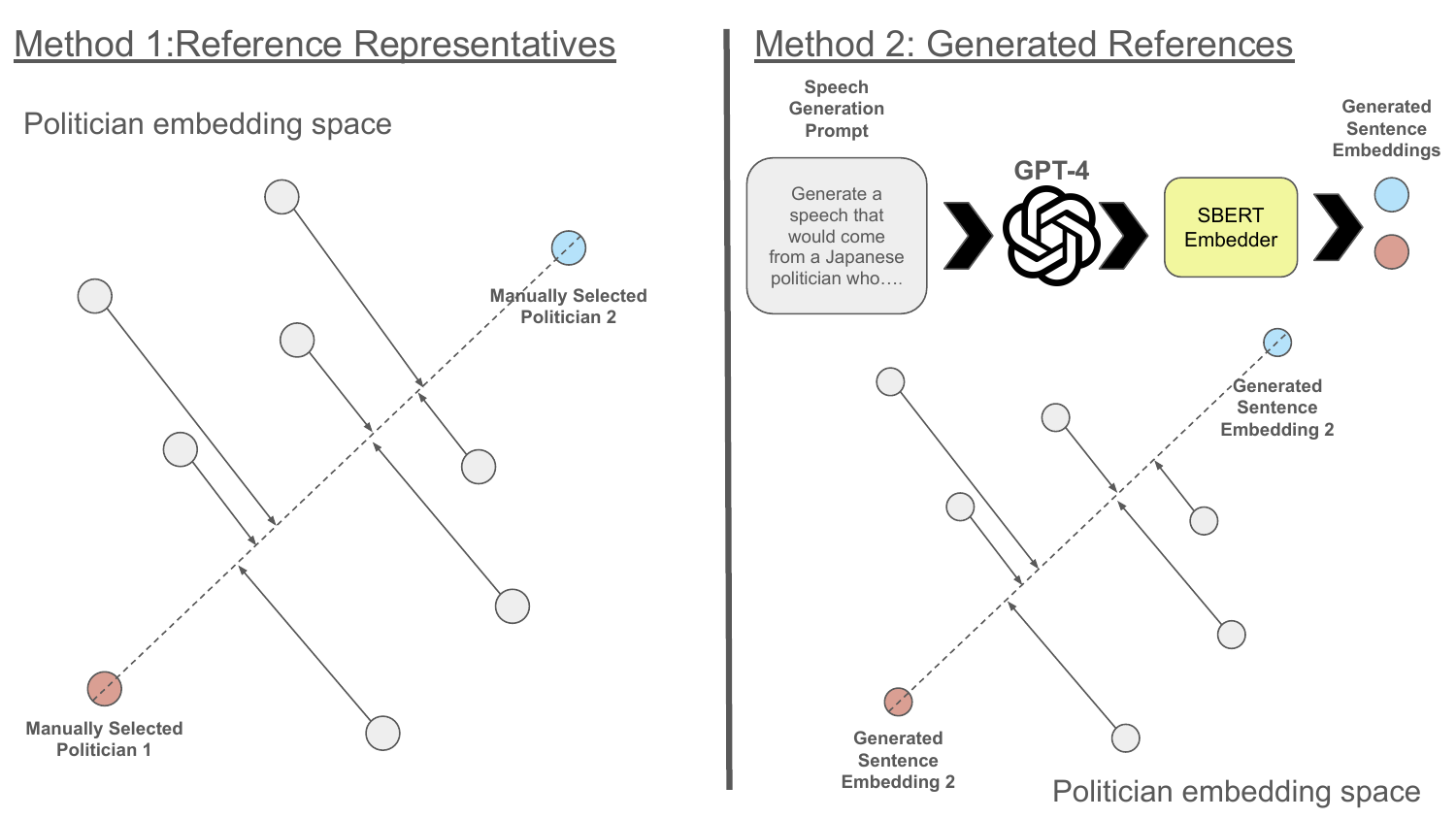}
  \caption{Graphical representation of the projection phase of the parliamentary speech embeddings}
  \label{fig:projectionphase}
\end{figure}

The methodology employed in this research involved a multi-step process for the analysis of parliamentary speeches which are graphically shown in figure \ref{fig:embeddingphase} and \ref{fig:projectionphase}. Initially, we collected representatives' speeches on targeted topics by employing various query words, subsequently refining the dataset by isolating opinion-based speech segments. To gauge the ideological stance of each politician on specific topics, we created an average opinion-based embedding for each representative for each topic. This involved utilizing the cl-tohoku/bert-base-japanese-v3\citep{Tohoku} model and the SBERT method.\citep{reimers2019sentencebert}

To collapse the embedding into a singular dimension, two distinct methods were pursued. In the first approach, we manually selected a representative pair known to hold opposing stances on the topic of interest. Subsequently, we projected all other opinion-based embedding onto the vector extending from one representative in the pair to the other. This method provided a scalar measurement of the ideological proximity of each representative to the reference points.

In the second method, we leveraged the GPT-4 model by OpenAI. We prompted it to generate 5 speeches that would come from a representative who are for/against the topic of interest(e.g. should Japan keep using nuclear power) and computed the average embedding of the generated sentences for each position. Then, similarly to the previous method, we projected all other opinion-based embedding for each representative onto the vector extending from one reference point to the other.

Additionally, an analysis of the latent space was conducted by reducing the dimensions of the embedding using the UMAP software. \citep{mcinnes2018umap-software} This allowed us to visualize the distribution of opinion-based embedding and generated sentence embedding on a 2D plane. By doing so, we gained insights into the spatial relationships between reference representatives, generated sentences, and the broader dataset, providing a comprehensive understanding of the ideological landscape within the Japanese parliamentary context.

And finally, we also conducted a qualitative semantic analysis by using the BERTopic method.\citep{grootendorst2022bertopic} We first split the projections of the opinion-based embeddings into three groups. These groups were then subjected to BERTopic which allowed us to conduct a detailed examination of the content discussed by different clusters within the dataset. This provided a nuanced understanding of the semantic variations among the groups, offering a qualitative validation of the model's ability to discern and categorize speeches based on underlying topics.

% \section{Validation}

% % \begin{itemize}
% %     \item project generated/picked sentence segments onto latent space and make sure that it aligns with our understanding of the dynamics in Japanese politics, similar to work done in \citep{Word-embeddings-for-analysis-of-ideological-placement}
% %     \item topic modelling on different groups on axis \citep{Word-embeddings-for-analysis-of-ideological-placement} following their pattern
% %     \item What are the different splitted groups saying in regard to different topics?
% %     - 
% % \end{itemize}
% The validation process in this research was conducted through two distinct methods, offering a comprehensive assessment of the reliability and relevance of the findings. In the first method, the positioning of the generated sentences in the latent space was analyzed using the UMAP method. This spatial analysis provided insights into the distribution of the sentences and how they align with the overall distribution of politicians from different parties. By mapping the generated sentences alongside the ideological landscape represented by the opinion-based embeddings, we were able to validate the consistency of the model's outputs with the known political affiliations of representatives.

\section{Results}
For this study, we explored two contentious issues in Japanese politics. Firstly, we looked at the discussions surrounding the necessity of conducting a public vote to determine whether the constitution should be amended to explicitly acknowledge the existence of the Japan Self-Defense Forces (JSDF). Following Japan's defeat in WWII, the Japanese constitution has explicitly stated that Japan cannot possess a military power which has become less aligned with reality over the past 80 years. The amendment to the constitution has been supported by right-leaning parties like the LDP and NDP, but opposed by left-leaning parties such as the Japanese Communist Party.

The second focal point revolves around the question of whether Japan should re-start nuclear power plants that are deemed safe to operate. Following the nuclear disaster of Fukushima in 2011, the Japanese public has been skeptical of the safety of the plants in an event of natural disaster. At the same time, Japan's over-reliance on coal and natural gas instead of clean-energy has also sparked discussions in the Japanese political world. Right-leaning parties have been supporting the decision to re-start plants that are deemed to be safe but left-leaning parties have been against it. 
\subsection{Projections}
Figure \ref{fig:Box JSDF} and \ref{fig:Box NPP} below show the box plots of politicians collapsed onto the references which are either manually picked or generated. For the topic of defence, we chose politician Tomomi Inada and Akira Kasai who we found to be for and against the acknowledgement of the JSDF in the constitution respectively after manual investigation. For the generated sentences, we prompted GPT-4 to generate a speech that would come from a politician who is for/against the acknowledgement of the JSDF in the Japanese constitution. 

Similarly, on the topic of nuclear power, we manually picked politician Fumio Kishida and Kazuo Shii who we found to be for and against the re-starting of the nuclear power plants in Japan respectively. We also prompted GPT-4 to generate speeches that would come from a politician who is for/against this point. The exact prompts used can be found in  \ref{app:gpt-prompts}.

\begin{figure}[h]
\centering
\begin{subfigure}{0.22\textwidth}
  \centering
  \includegraphics[width=1\linewidth]{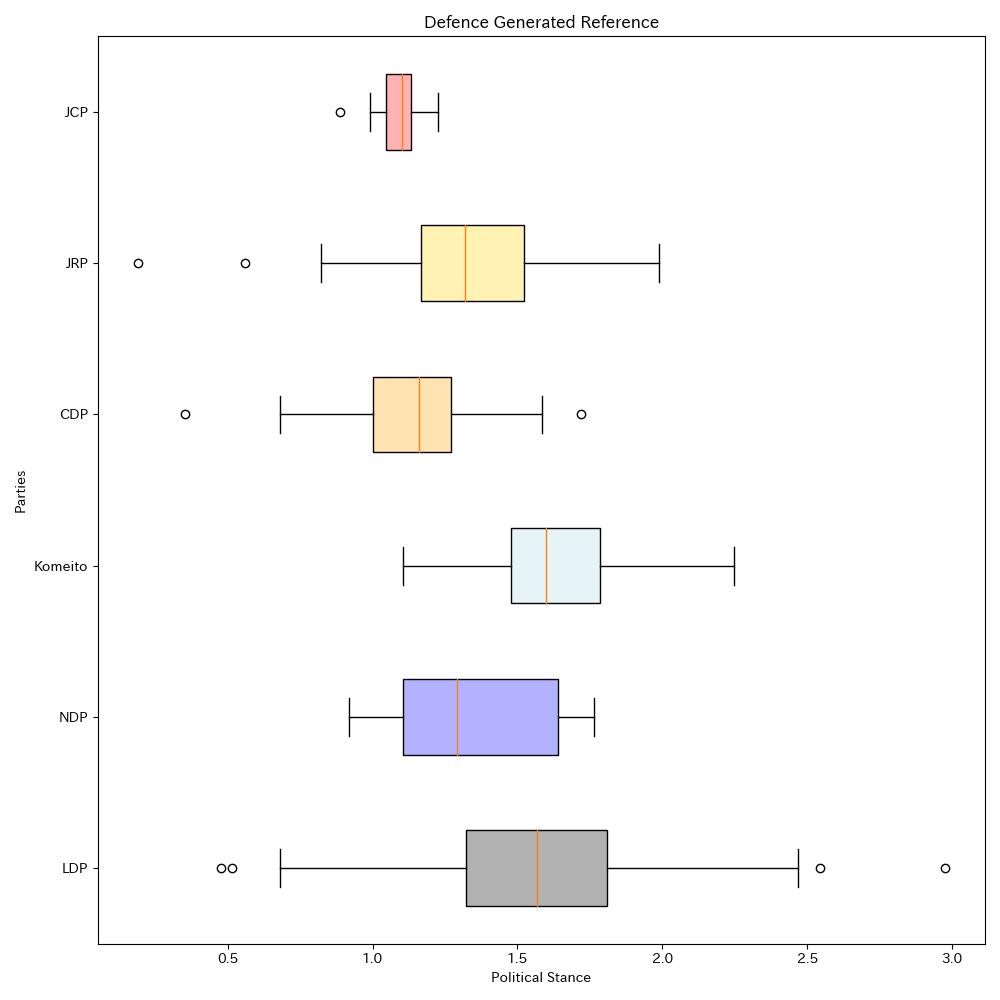}
  \caption{Defence collapsed on generated reference points}
  \label{fig:sub1}
\end{subfigure}
\begin{subfigure}{0.22\textwidth}
  \centering
  \includegraphics[width=1\linewidth]{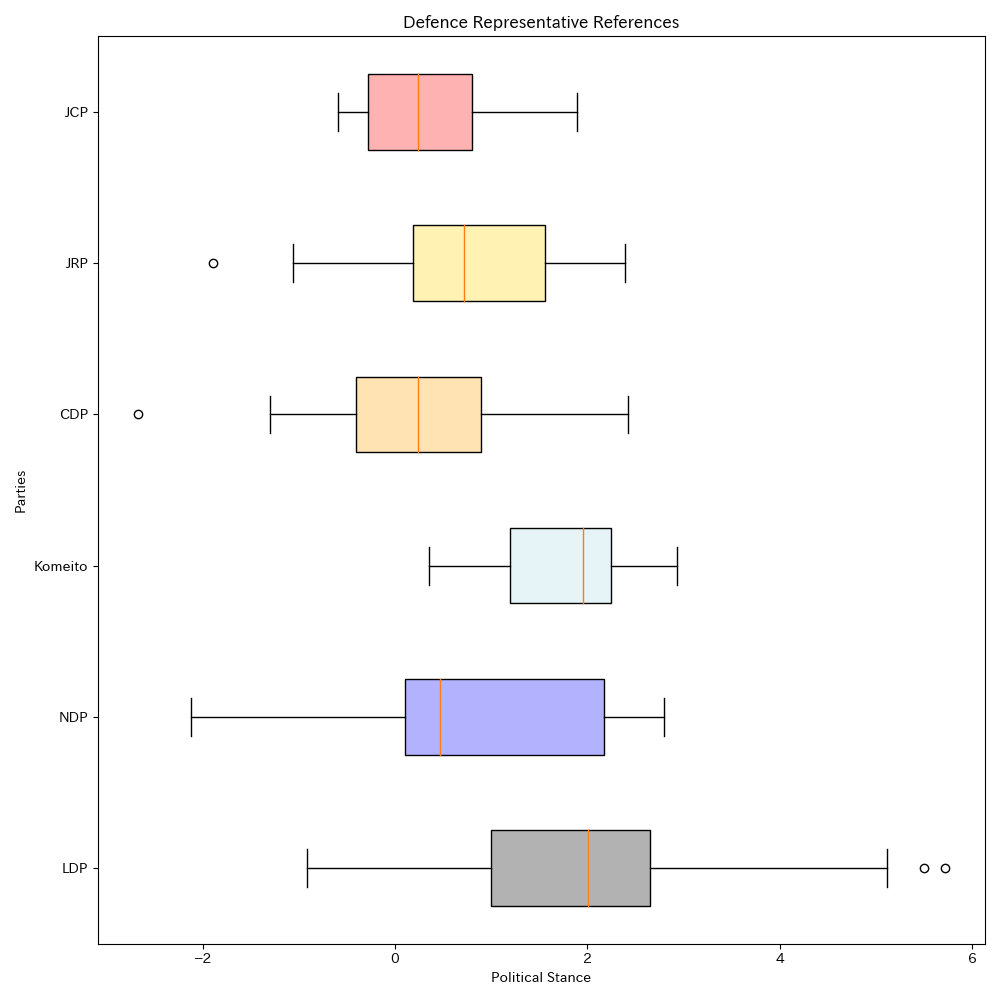}
  \caption{Defence collapsed on reference politicians}
  \label{fig:sub2}
\end{subfigure}
\caption{Representatives positioned on a pro-contra scale in regards to acknowledgement of the JSDF in the Japanese Constitution}
\label{fig:Box JSDF}
\end{figure}

\begin{figure}[h]
\centering
\begin{subfigure}{0.22\textwidth}
  \centering
  \includegraphics[width=1\linewidth]{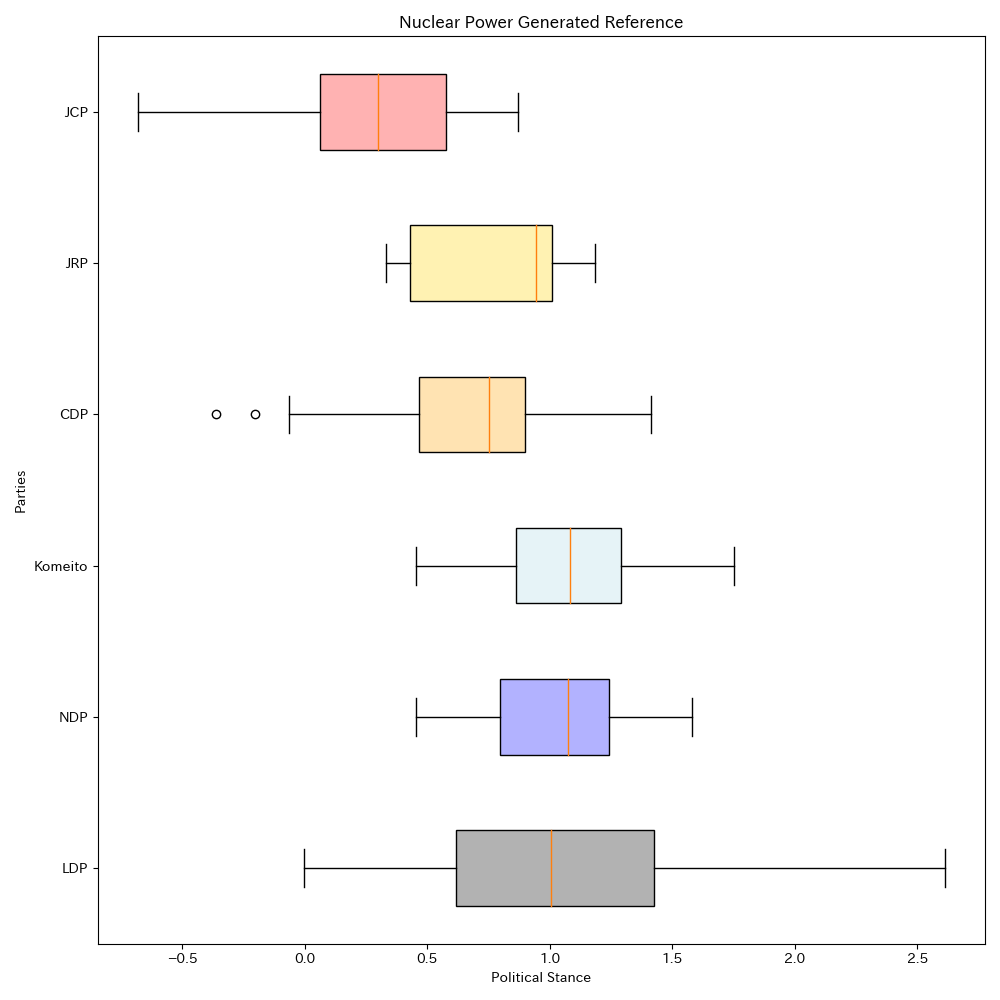}
  \caption{Nuclear collapsed on generated reference points}
  \label{fig:sub1}
\end{subfigure}
\begin{subfigure}{0.22\textwidth}
  \centering
  \includegraphics[width=1\linewidth]{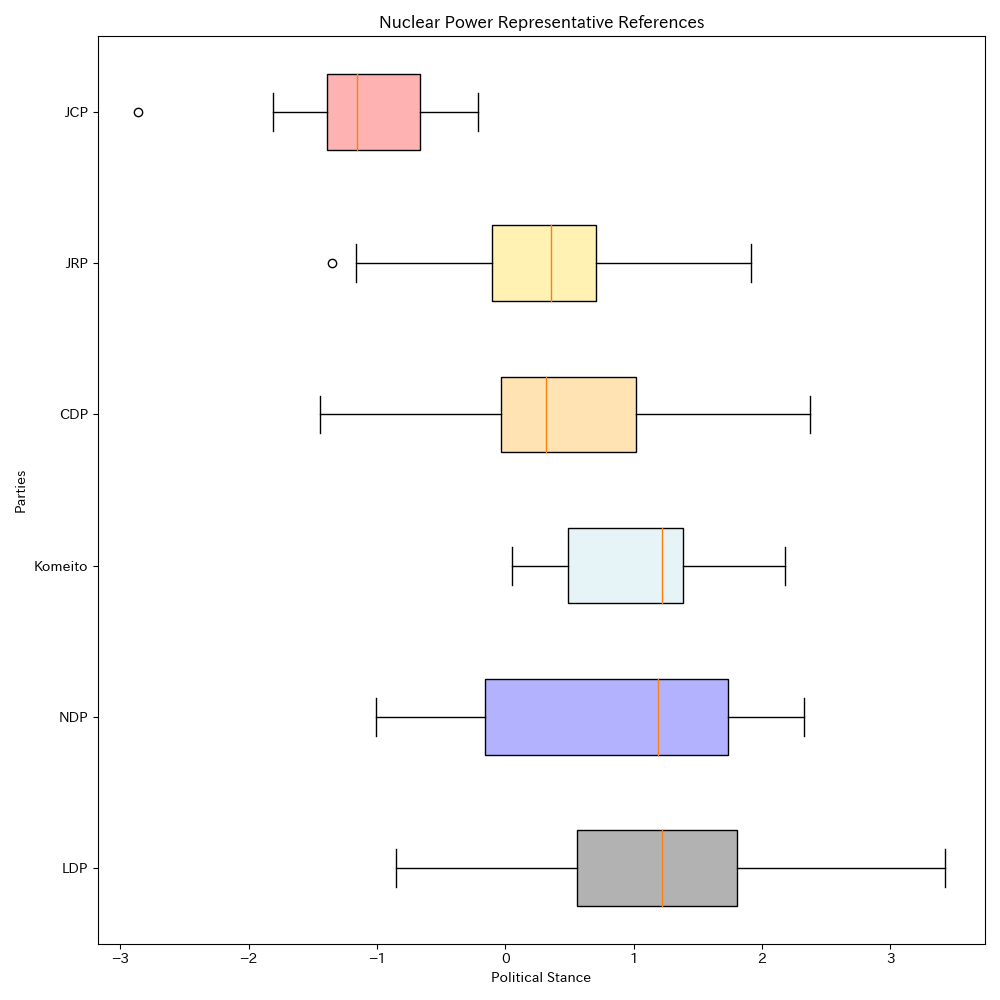}
  \caption{Nuclear collapsed on reference politicians}
  \label{fig:sub2}
\end{subfigure}
\caption{Representatives positioned on a pro-contra scale in regards to restarting the NPPs in Japan}
\label{fig:Box NPP}
\end{figure}

\subsection{UMAP analysis}
\begin{figure}[h]
\centering
\begin{subfigure}{0.22\textwidth}
  \centering
  \includegraphics[width=1\linewidth]{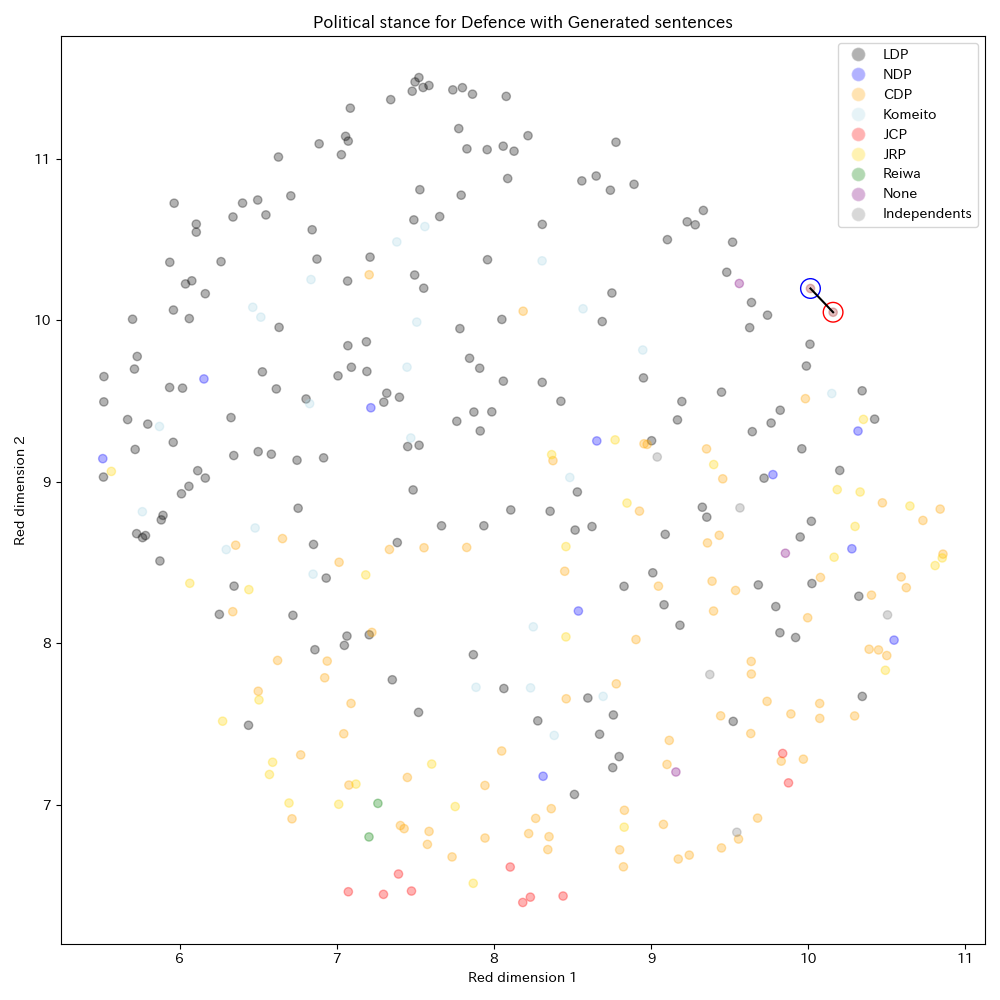}
  \caption{Defence collapsed on generated reference points}
  \label{fig:sub1}
\end{subfigure}
\begin{subfigure}{0.22\textwidth}
  \centering
  \includegraphics[width=1\linewidth]{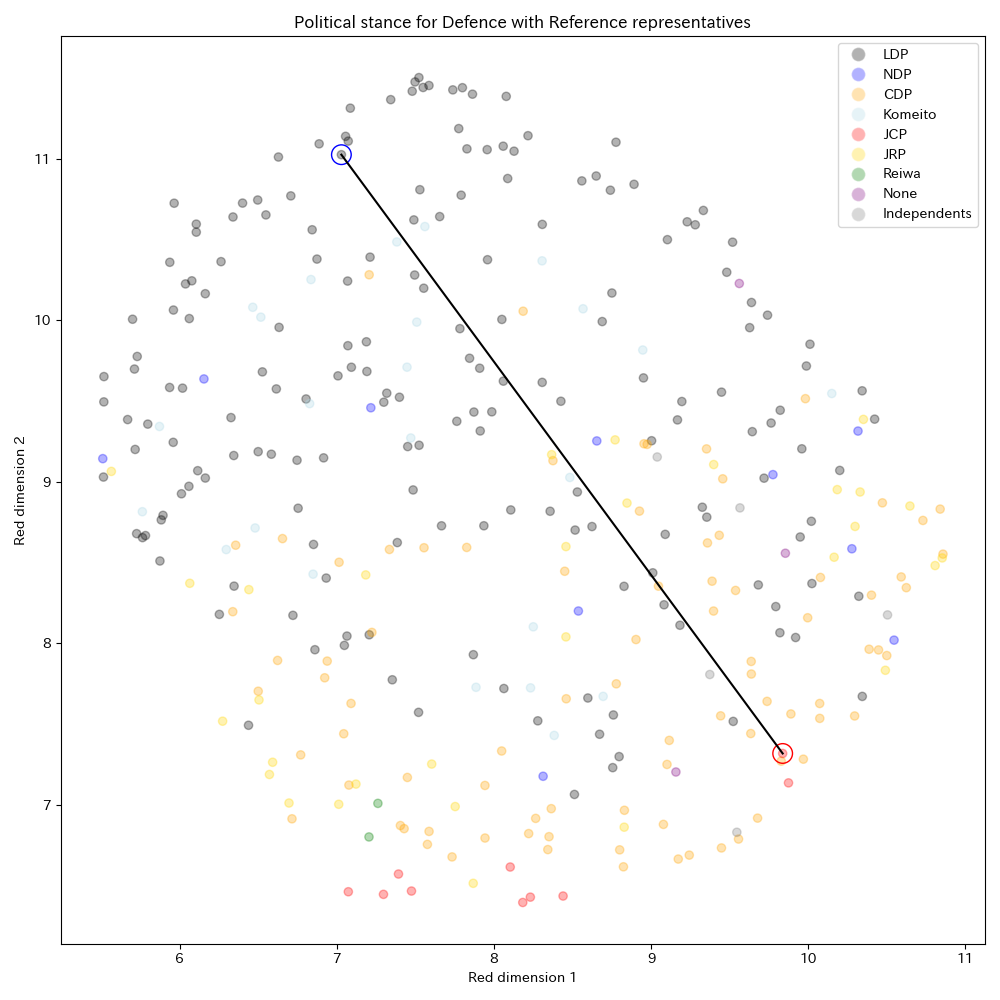}
  \caption{Defence collapsed on reference politicians}
  \label{fig:sub2}
\end{subfigure}
\caption{UMAP plot of the opinion embeddings of representatives on the topic of acknowledgement of JSDF in the constitution}
\label{fig:UMAP JSDF}
\end{figure}

\begin{figure}[h]
\centering
\begin{subfigure}{0.22\textwidth}
  \centering
  \includegraphics[width=1\linewidth]{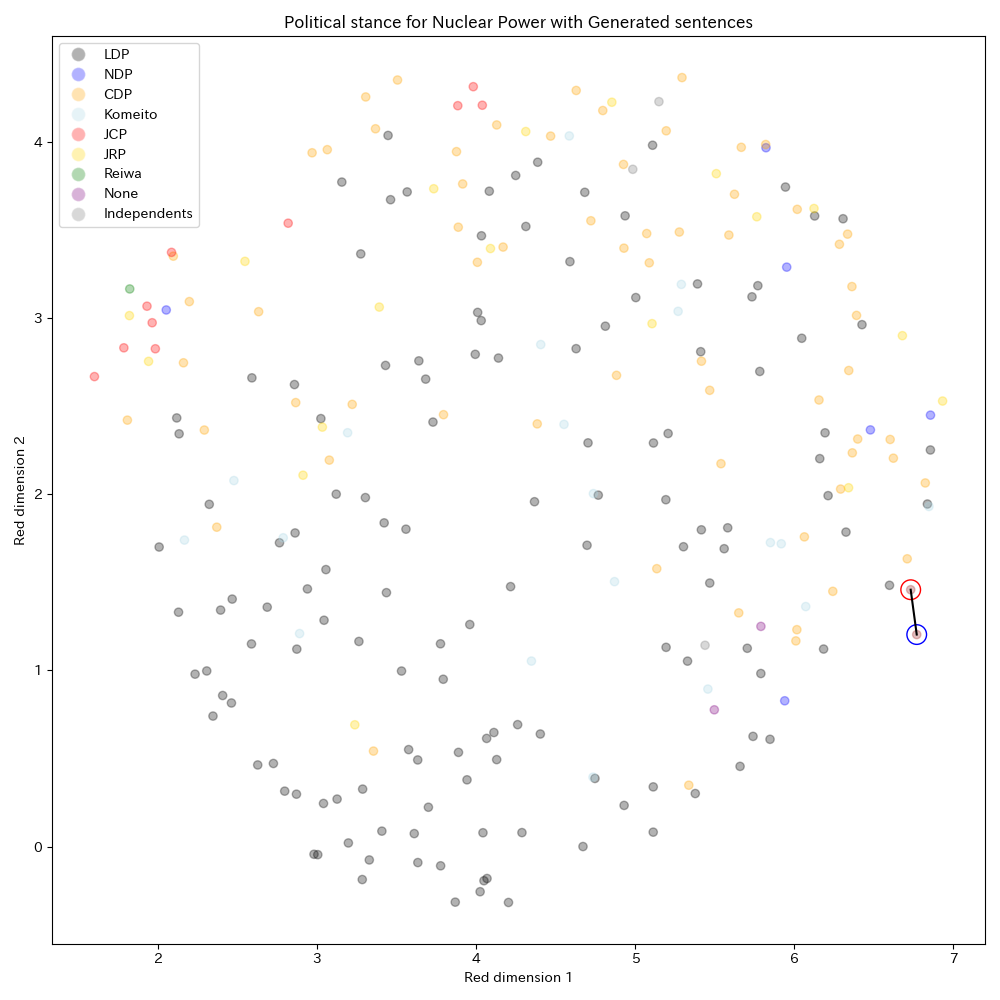}
  \caption{Nuclear collapsed on generated reference points}
  \label{fig:sub1}
\end{subfigure}
\begin{subfigure}{0.22\textwidth}
  \centering
  \includegraphics[width=1\linewidth]{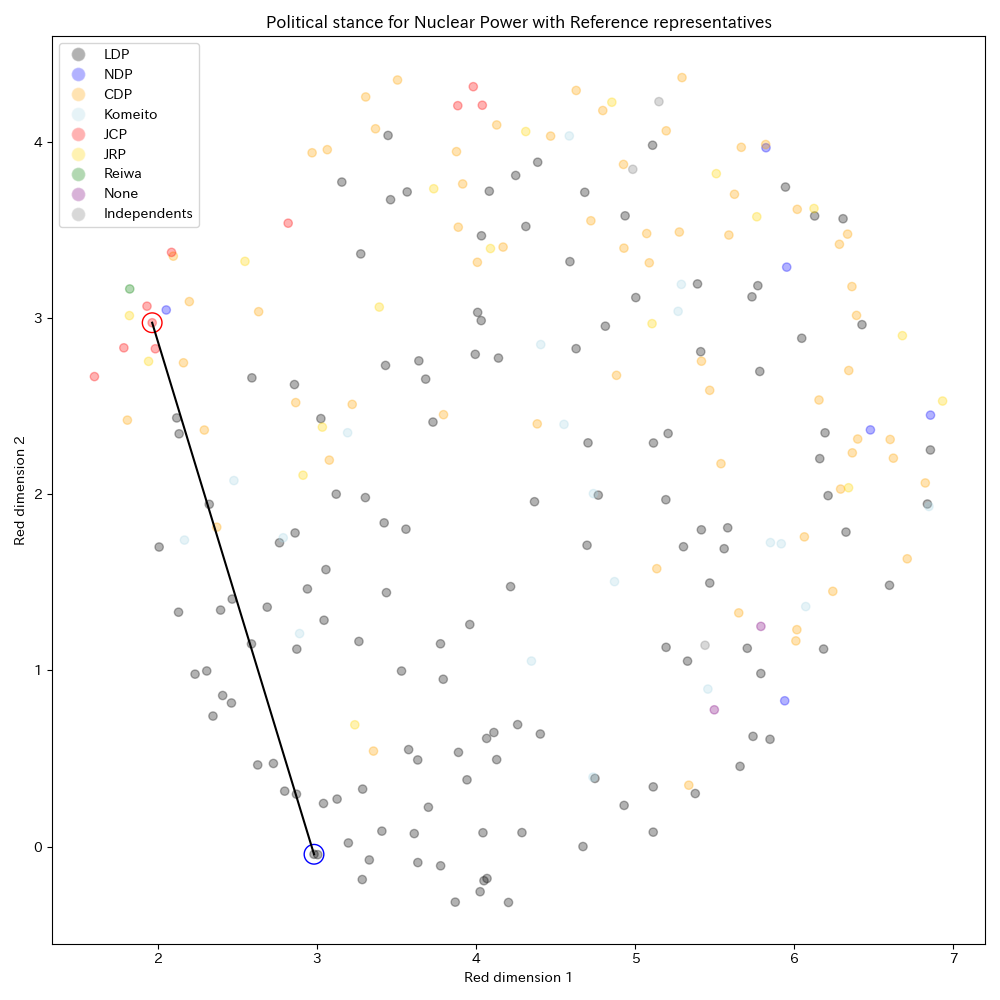}
  \caption{Nuclear collapsed on reference politicians}
  \label{fig:sub2}
\end{subfigure}
\caption{UMAP plot of the opinion embeddings of representatives on the topic of restarting of NPPs}
\label{fig:UMAP NPP}
\end{figure}

See figure \ref{fig:UMAP JSDF} and \ref{fig:UMAP NPP} for lower the dimension visualizations of the opinion-embeddings. We also have visualized the location of the references with larger circles colored blue and red representing the reference points for and against the topic of interest respectively with a black line joining the reference points. 

On the plots, we notice that the lines connecting the representative references and the generated references are parallel to each other, indicating that along the displayed line, politician stances change from favourable to unfavourable in regards to the respective topics.

More specifically, we see that for the topic of defence the bottom right segment of the plot are representatives who are against the acknowledgement of the JSDF in the constitution and the top left segment are representatives who are for the acknowledgement. Similarly for the topic of nuclear power, we see that politicians who are against the use of nuclear power are on the upper side of the plot and politicians who are for it are on the bottom side.

% #######################BERTOPIC#####################################
\subsection{BERTopic results}
In this section we present the interpretations of the BERTopic results we applied on the 3 group segments which are summarized in Table \ref{tab:Bertopic-defence} and \ref{tab:Bertopic-nuclear}. Here we point out some noticeable observations. The raw Japanese BERTopic results are shown in tables \ref{tab:Bertopic defence raw} and \ref{tab:Bertopic nuclear raw} in the appendix.
\begin{itemize}
    \item Acknowledgement of JSDF in constitution
    \begin{itemize}
        \item We see that the group that is against this topic emphasize US bases in Okinawa. This is a common topic that is raised by the group as parties such as the JCP have been advocating for the removal of US bases from Japanese soil.
        \item We observe that the group that is favourable of this mentions the risk that surrounds Japan by mentioning the Taiwan strait crisis and the Russian invasion of Ukraine. 
    \end{itemize}
    \item Restarting of NPPs
    \begin{itemize}
        \item We notice that the group against the restart of NPPs emphasize the safety and responsibility of operating NPPs as well as the accident. 
        \item While the group favourable of the restart also talk about the accident and the risks associated with the NPPs, they shift their attention more towards moving forward from the accident 
    \end{itemize}
\end{itemize}

\begin{table}[htbp] % or other placement options such as [!htb]
\centering
\caption{English Interpretation of BERTopic Results for Defence}\label{tab:Bertopic-defence}
\begin{tabularx}{.45\textwidth}{c|X|X|X } 
\hline
Topics  & Group 0  & Group 1  & Group 2 \\
& (Left) & (Centered) & (Right)\\
\hline
Topic 0 & Japanese Government's relationship with the US & Japanese government and people & US-Japan Relationship and Diplomacy \\ 
\hline
Topic 1 & US-bases in Okinawa & Issues surrounding US deployment in Japan & Taiwan Strait Risk  \\ 
\hline
Topic 2 & Parties in Japanese Diet & International Condition Surrounding Japan & Accidents caused by helicopters of JSDF \\ 
\hline
Topic 3 & Self-sustainability of Japan and Diplomacy & Funding of JSDF  & Ukraine Aid and Natural disaster aid \\ 
\hline
Topic 4 & Increased funding of JSDF & Japanese law and limits of JSDF activities & Diplomacy and Military Deterrent\\
\hline
\end{tabularx}
\end{table}

\begin{table}[htbp] % or other placement options such as [!htb]
\centering
\caption{English Interpretation of BERTopic Results for Nuclear Power}\label{tab:Bertopic-nuclear}
\begin{tabularx}{.45\textwidth}{c|X|X|X } 
\hline
Topics  & Group 0  & Group 1  & Group 2 \\
& (Left) & (Centered) & (Right)\\
\hline
Topic 0 & Accident of Power Plant & Accident and Electricity Supply & Accident and Energy \\ 
\hline
Topic 1 & Fukushima Accident and Responsibilities &  Damages caused by Fukushima accident & Damages of Fukushima Disaster \\ 
\hline
Topic 2 & Government and Enterprises & Accident, Chenobyl and the learnings  & Fukushima Disaster and Challenges\\ 
\hline
Topic 3 & Assurance of Safety & 3.11 Earthquake and Responses & Local area of Fukushima and long-term solution \\ 
\hline
Topic 4 & (Un-interpretable) & (Un-interpretable) & Russian Invasion of Ukraine and attack on Zaporizhzhya NPP\\
\hline
\end{tabularx}
\end{table}

\section{Comparison to the estimations of Mielka}
\begin{figure}[h]
\centering
  \centering
  \includegraphics[width=0.8\linewidth]{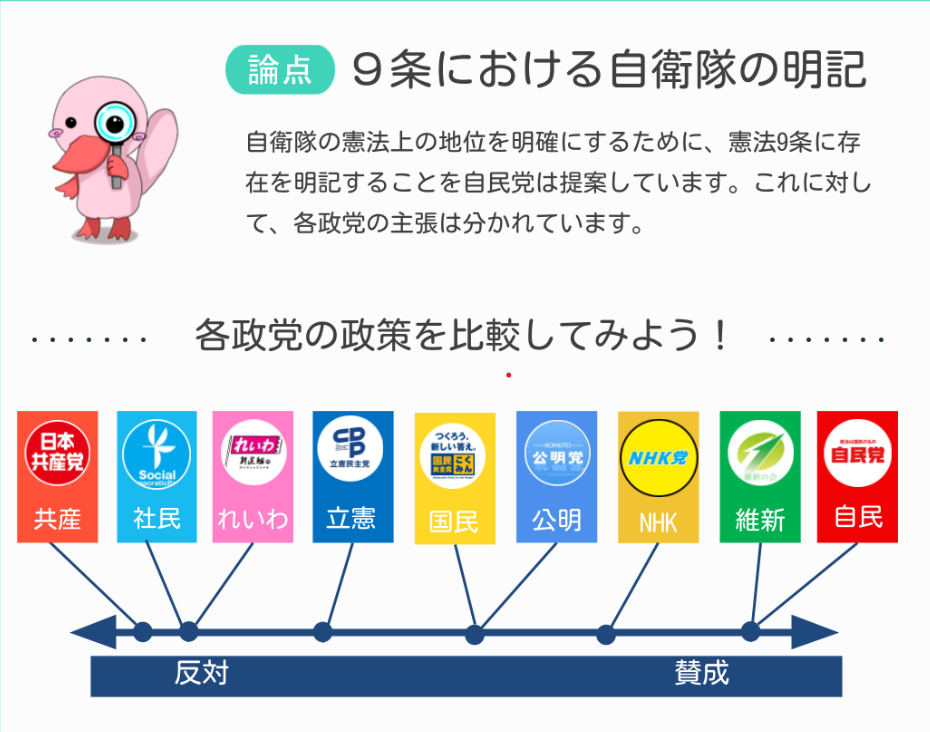}
  \caption{Policy Position Estimations By Mielka on Acknowledgement of JSDF in Constitution \citep{Mielka}}
  \label{fig:Mielka_defence}
\end{figure}

\begin{figure}[h]
\centering
  \centering
  \includegraphics[width=0.8\linewidth]{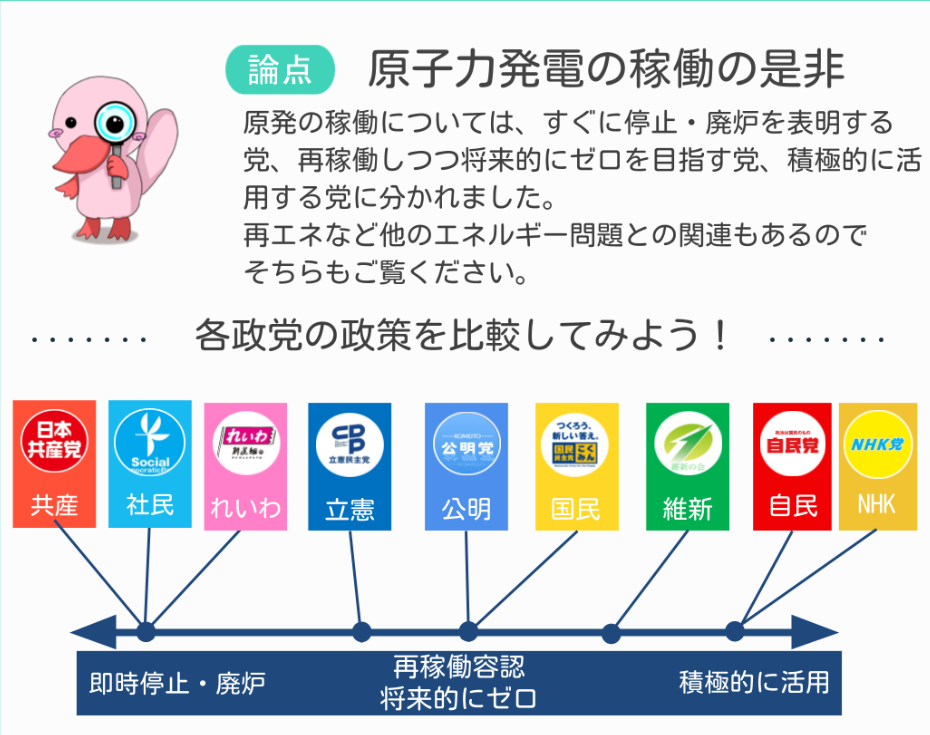}
  \caption{Policy Position Estimations by Mielka on Restarting NPPs \citep{Mielka}}
  \label{fig:Mielka_nuclear}
\end{figure}

In this section, we will be discussing the results obtained by our analysis by comparing it to the estimations of Mielka (see Figure \ref{fig:Mielka_defence} and \ref{fig:Mielka_nuclear} for the policy position estimations of the parties by Mielka\citep{Mielka}. In the process of this research, we reached out to the organization to inquire about their methodology of estimation. They replied that this is done by 3 experts discussing the policy position of the parties by looking at party manifestos and policy summaries and positioning the party upon mutual agreement of the estimation. Note that while their estimation includes all the parties, we have focused on the 6 major parties in Japan. Comparing the estimations of Mielka to our estimations, there are a few observations to be made. Below we summarize these observations for each of the topics. 

\subsection{Comparison of Policy Positions - JSDF}
\begin{itemize}
    \item \textbf{Our estimations capture the overall positions of the parties. }
        \begin{itemize}
            \item The JCP and CDP who are estimated by Mielka to be against the acknowledgement of JSDF are also positioned on the left side by our estimation.
            \item The LDP, Komeito, JRP, and NDP who are estimated by Mielka to be relatively pro-acknowledgement of JSDF are positioned on the right side by our estimation. 
        \end{itemize}
\end{itemize}

\subsection{Comparison of Policy Positions - NPP}
\begin{itemize}
    \item \textbf{Our estimations capture the overall positions of the parties. }
        \begin{itemize}
            \item JCP and CDP who are estimated by Mielka to be against the re-starting of NPPs are positioned on the left side as well by our estimation. 
            \item Komeito, JRP, and NDP who are estimated to be relatively neutral in regards to restart of NPPs are positioned centrally by our estimation.
            \item LDP who are known to be favourable of restarting the NPPs are placed to the right. 
        \end{itemize}
\end{itemize}

\section{Limitation of our findings}
In this section, we discuss some limitations of our methodology and validation process. 
\subsection{Representing each legislator as a single embedding}
In our methodology, we are representing each representative as a single embedding by taking the average of the embeddings of the opinion-based sentences spoken by the representative in the past. Therefore in this methodology, if a representative changes their political stance suddenly, their stance-embedding is going to be skewed by their speeches from the past. This could be improved by weighting the more recent speeches more compared to the older speeches but this is something to be explored by future research.  

\subsection{Discrepancy to the estimations of Mielka}
By comparing the estimations by Mielka to our estimations, we notice that they do not mirror each other perfectly despite being generally similar in terms of party positions. Here we discuss possible explanations for this discrepancy. 

\subsubsection{Parties paint a specific picture about themselves}
We notice that there are some discrepancies on the relative positions of the parties between our estimations and the estimations by Mielka. For example for the topic of acknowledgement of the JSDF, while Komeito is positioned more centrally by Mielka, by our estimation it is positioned more pro-acknowledgement. This discrepancy suggests that some parties are advertising themselves in a certain way through mediums which are more visible to the public (e.g., party manifestos, policy summaries) which might not necessary be aligned with their utterances in the diet. Since the estimations of Mielka are based on the party manifestos and policy summaries published by the parties themselves, their estimations might not accurately reflect the stances taken by the representatives in the diet. 

\subsubsection{Legislators are not created equally}
Members of parties do not have an equal say in the manifestos and policies published by the parties. Parties follow a strict hierarchical structure based on the number of elections of representatives and the positions held by members. However, our box plots do not take this into account and treat all representative equally. Since the Mielka estimations are based on the party manifestos, it is possible that their estimations are skewed towards representatives with more influence in the party.

\subsection{LLMs are black-boxes}
 We do not know the text corpora that was used to train the LLM and the bias introduced during the training process. This is an inevitable issue that comes with using a third-party pre-trained LLM in any research. However during our research, we have made the effort to visualize the embeddings of the LLM and interpret how the LLM is embedding the documents. We believe that through this effort, we were able to present the embeddings in a human-understandable manner.

\newpage
\section{Conclusion}
Methods leveraging the latent knowledge of LLMs are most likely going to be the main methodology going forward to analyze political text due to their accessibility and reliability for political scientists. This paper showcased that we are able to retrieve political ideological scaling purely from the utterances of individual politicians in the parliament that align with pre-existing expert predictions. This makes the research of political text significantly easier for researchers as they no longer need to fit their own model to political text which will require computational power and high level programming knowledge. 

In the future, we see this methodology being extended in two directions. The first is to analyze the change in ideological stance of politicians over-time, similar to the works of \citeauthor{Word-embeddings-for-analysis-of-ideological-placement} and \citeauthor{gentzkow-measuring-group-differences}.  Being able to quantitatively express the shift in political ideology is going to be a valuable measure for the public and researchers. 

In this paper, we have only shown our methodology in the context of Japanese politics. But applying and verifying our result in other parliamentary systems would help validate our methodology further.

\appendix
\section{Query words}

\begin{table}[htbp] % or other placement options such as [!htb]
\centering
\caption{English translations of query words}\label{tab:English-query}
\begin{tabularx}{.45\textwidth}{ c|X } 
\hline
Topic & Query Words \\  
\hline
Defence & JSDF, Defence, Security, Collective Self-Defence, Military Power \\ 
\hline
Nuclear Power & Nuclear, Nuclear Power Plant, Decommissioning, Restarting, Baseload \\
\hline
\end{tabularx}
\end{table}

\begin{CJK}{UTF8}{min}
\begin{table}[htbp] % or other placement options such as [!htb]
\centering
\caption{Japanese query words}\label{tab:Japanese-query}
\begin{tabularx}{.45\textwidth}{ c|X } 
\hline
 Topic & Query Words \\  
 \hline
  防衛 & 自衛隊,集団的自衛権,安全保障,軍事力,防衛 \\
  \hline
  原発 & 原子,原発, 廃炉,再稼働,ベースロード  \\
 \hline
\end{tabularx}
\end{table}
\end{CJK}
% ##################GPT prompts############################
\section{GPT-prompts}
\label{app:gpt-prompts}
\begin{quote}
    \begin{CJK}{UTF8}{min}
    自衛隊を憲法に明記することに賛成な政治家の答弁を作ってください。\newline
    \textbf{English Translation: }Please write a response from a politician who is in favor of acknowledging the Self-Defense Forces in the Constitution.
    \end{CJK}
\end{quote}

\begin{quote}
    \begin{CJK}{UTF8}{min}
        原発の再稼働に賛成な政治家の答弁を作ってください。\newline
         \textbf{English Translation: }Please write a response from a politician who is in favor of restarting nuclear power plants.
    \end{CJK}
\end{quote}

% ##################PARTY NAMES TABLE############################
\newpage
\section{Party Names and Abbreviations}
\begin{CJK}{UTF8}{min}
\begin{table}[htbp] % or other placement options such as [!htb]
\centering
\caption{Party names and abbreviations}\label{tab:Parties and abbrev}
\begin{tabularx}{.45\textwidth}{ c|X|X } 
\hline
 Party Name(JP) & Party Name(EN) & Abbreviation\\  
 \hline
  自由民主党  & Liberal Democratic Party & LDP \\
  \hline
  国民民主党  & National Democratic Party & NDP \\
  \hline
  立憲民主党  & Constitutional Democratic Party & CDP \\
  \hline
  日本共産党  & Japanese Communist Party & JCP \\
  \hline
  公明党  & Komeito & Komeito \\
   \hline
  日本維新の会  & Japan Restoration Party & JRP \\
  \hline
\end{tabularx}
\end{table}
\end{CJK}

% ##################BERTOPIC RESULTS############################
\clearpage
\section{Raw BERTopic results}

\begin{CJK}{UTF8}{min}
\begin{table}[htbp] % or other placement options such as [!htb]
\centering
\caption{BERTopic Results for Defence}
\label{tab:Bertopic defence raw}
\begin{tabularx}{.45\textwidth}{c|X|X|X } 
\hline
Topics  & Group 0  & Group 1  & Group 2 \\
& (Left) & (Centered) & (Right)\\
\hline
Topic 0 & 'わけ', '国', '大臣', '法', '政府', 'アメリカ', '庁', '国民', '場合', '話'& '日本', '国', '中', '国民', '大臣', '問題', '御', '庁', '状況', '性' & '国', '大臣', '中', '省', '環境', '御', '国際', '国務', '国民', '日米' \\ 
\hline
Topic 1 & '国', '理事', '沖縄', '早期', '環境', '不', '体制', '等', '改革', '性'& '法', '日本', '問題', '自民', '案', 'アメリカ', '基地', '米軍', '項', '大臣' & '台湾', '社会', '海峡', '国際', '挑発', '全体', '脅威', '国', '行為', '一連' \\ 
\hline
Topic 2 & '党', '自民', '政権', '維新', '民主', '民進', '会', '立場', '安倍', '法'& '環境', '国', '課題', '同盟', '国際', '地域', '体制', '中', '化', 'アジア' & '事故', '再発', '存立', '住民', '職責', '省', '原因', '重み', '墜落', 'ヘリコプター' \\ 
\hline
Topic 3 & '食料', '自給', '政策', '責任', '前進', '大臣', '率', '外交', '王道', '性'& '財源', '省', '年度', '御', '庁', '国民', '予算', '円', '施設', '収入' & '所存', '課題', '全力', '省', '災害', 'ウクライナ', '救援', '大臣', '援助', 'コロナ' \\ 
\hline
Topic 4 & '費', '財源', '％', '予算', '剰余', '決算', '円', '増', '歳出', '予備'& '場合', '国', '要件', '法', '最小', '目的', '第', '大臣', '限度', '行為' & '能動', '分野', '外交', '未然', '国力', '危機', '交渉', '最大', '国際', '帰属'\\
\hline
\end{tabularx}
\end{table}
\end{CJK}

\begin{CJK}{UTF8}{min}
\begin{table}[htbp] % or other placement options such as [!htb]
\centering
\caption{BERTopic Results for Nuclear Power}
\label{tab:Bertopic nuclear raw}
\begin{tabularx}{.45\textwidth}{c|X|X|X } 
\hline
Topics  & Group 0  & Group 1  & Group 2 \\
& (Left) & (Centered) & (Right)\\
\hline
Topic 0 & '問題', '中', '力', '本当', 'ふう', 'わけ', '事故', '話', '大臣', '法' & '力', '事故', 'ふう', 'わけ', '日本', 'エネルギー', '国', '性', '政策', '電力' & 'エネルギー', '御', '中', '大臣', '性', 'わけ', '国', '問題', '事故', '国務' \\ 
\hline
Topic 1 & '事故', 'エネルギー', '力', '国民', '第', '国', '大', '責任', '福島', '東京'& '力', '者', '事故', '国', '事業', '損害', '会', '政府', '性', '法' & '等', '国', '者', '会', '性', '事故', '損害', '研究', '所', '政策' \\ 
\hline
Topic 2 & '声', '中', '世界', '政府', '極論', '基', '頭', '経営', '2', '企業'& '事故', '力', '人類', 'パネル', '国家', 'チェルノブイリ', '工程', '日本', '教訓', 'イギリス' & '福島', '最', '第', '経済', '水', '課題', '事故', '産業', '全力', '所'\\ 
\hline
Topic 3 & '長', '力', '田', '更', '当時', '現実', '機構', '神話', '会', '保安'& '第', '事故', '震災', '水', '東', '日本', '電力', '東京', '取扱い', '対策' & '地域', '災害', '福島', '中長期', '者', '本格', '事故', '方々', '被害', '県' \\ 
\hline
Topic 4 &'法', '案', '一気', '前回', '回帰', 'イエス', 'はず', '与党', 'ノー', '審議' & '中長期', '方向', '対立', '項', '最大', '脱', '性', '力', '基本', '過半' & 'ロシア', '攻撃', 'ウクライナ', 'ザポリージャ', '即座', '行為', '施設', '暴挙', '一連', '所'\\
\hline
\end{tabularx}
\end{table}
\end{CJK}
\clearpage
\bibliographystyle{elsarticle-harv} 
\bibliography{references}
\end{document}